\documentclass[runningheads]{llncs}
\usepackage[T1]{fontenc}
\usepackage{graphicx}
\usepackage{amsmath,amsfonts,amssymb}
\usepackage{graphicx}
\usepackage{multirow}
\usepackage[colorlinks=true, allcolors=blue]{hyperref}
\begin{document}
\title{USegMix: Unsupervised Segment Mix for Efficient Data Augmentation in Pathology Images}
\titlerunning{Unsupervised Segment Mix for Efficient Data Augmentation}
%
%

\author{Jiamu Wang \and JinTae Kwak}
\authorrunning{J. Wang \& J.T. Kwak}
\institute{School of Electrical Engineering, Korea University, Seoul 02841, Korea\\
\email{\{taurusmumu,jkwak\}@korea.ac.kr}}
\maketitle              
\begin{abstract}
In computational pathology, researchers often face challenges due to the scarcity of labeled pathology datasets. Data augmentation emerges as a crucial technique to mitigate this limitation. In this study, we introduce an efficient data augmentation method for pathology images, called USegMix. Given a set of pathology images, the proposed method generates a new, synthetic image in two phases. In the first phase, USegMix constructs a pool of tissue segments in an automated and unsupervised manner using superpixels and the Segment Anything Model (SAM). In the second phase, USegMix selects a candidate segment in a target image, replaces it with a similar segment from the segment pool, and blends them by using a pre-trained diffusion model. In this way, USegMix can generate diverse and realistic pathology images. We rigorously evaluate the effectiveness of USegMix on two pathology image datasets of colorectal and prostate cancers. The results demonstrate improvements in cancer classification performance, underscoring the substantial potential of USegMix for pathology image analysis.

\keywords{Data Augmentation \and Cancer Classification \and Unsupervised Learning \and Semantic Segmentation \and Diffusion Model Inpainting.}
\end{abstract}

\begin{figure}
\centering
\includegraphics[width=\linewidth]{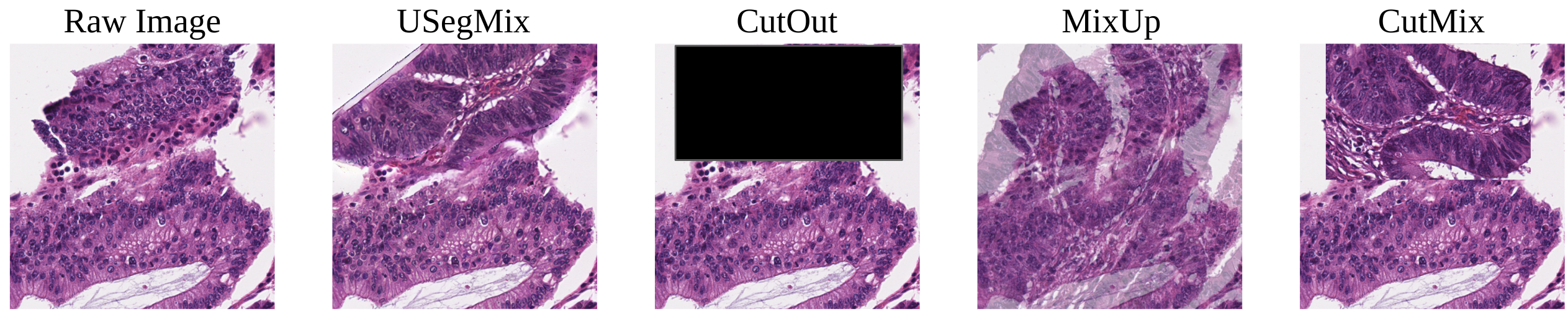}
\caption{Comparison among USegMix(ours) and CutOut, MixUp, CutMix.} \label{fig0}
\end{figure}

\section{Introduction}
Computational pathology is a transformative approach in pathology research and practice, utilizing advanced computational techniques such as machine learning, deep learning, and digital imaging to improve disease diagnosis and management~\cite{medi_aug}. Despite its potential, it faces challenges such as the scarcity of annotated pathology datasets. Reliance on expert annotations and ethical considerations regarding patient consent limits the availability of diverse data, affecting the generalizability of computational models and hindering their application in real-world clinical settings.

To address data limitations in pathology, data augmentation has become essential for enhancing dataset size and diversity. Early methods used simple manipulations like image flipping, rotation, and cropping to introduce variability. Later, more advanced mix-based techniques emerged; for instance, CutOut~\cite{cutout} simulates dropout by masking parts of the image, MixUp~\cite{mixup} blends two images in the alpha channel, and CutMix~\cite{cutmix} replaces a region in one image with content from another to create a composite image. Style Transfer~\cite{style_trans} merges the content of one image with the style of another, preserving content while adding stylistic variations. More recently, Attentive CutMix~\cite{att_cutmix} replaces the most descriptive area according to the attention maps, TransMix~\cite{transmix} mixes labels based on the attention maps of Vision Transformers,  ClassMix~\cite{classmix}, designed for semantic segmentation, mixes samples based on segmentation boundaries, CarveMix~\cite{carvemix} creates new brain lesion images by copying and pasting lesions, GradMix\cite{gradmix} and InsMix~\cite{insmix} increase the diversity of nuclei by employing the copy-and-paste approach. These methods aim to provide richer, more varied training samples, overcoming traditional augmentation limitations. However, it is likely that such techniques introduce artifacts and thus create unrealistic images, which is a critical issue in computational pathology, as shown in Fig.~\ref{fig0}.

In this study, we introduce USegMix, an efficient augmentation method designed to enhance the realism and diversity of pathology image datasets. USegMix creates a new, synthetic image in two phases. First, it decomposes pathology images into a set of segments in an unsupervised manner using superpixels and the Segment Anything Model (SAM)~\cite{sam}, further constructing a tissue segment pool. Second, it selects a candidate tissue segment in a target image, replaces the candidate segment with a semantically similar segment, belonging to the same histologic category, from the tissue segment pool, and seamlessly blends the target image and the selected segment using a pre-trained diffusion model~\cite{dm,ddpm,smld,sde,ldm}. In this manner, USegMix can generate a variety of synthetic images that are realistic and semantically consistent without additional training on a target dataset. We evaluated USegMix using two public cancer datasets. The results demonstrate that USegMix increases the size and diversity of pathology datasets and improves the performance of downstream tasks, highlighting its potential to enhance computational pathology through advanced data augmentation.

\begin{figure}
\centering
\includegraphics[width=\linewidth]{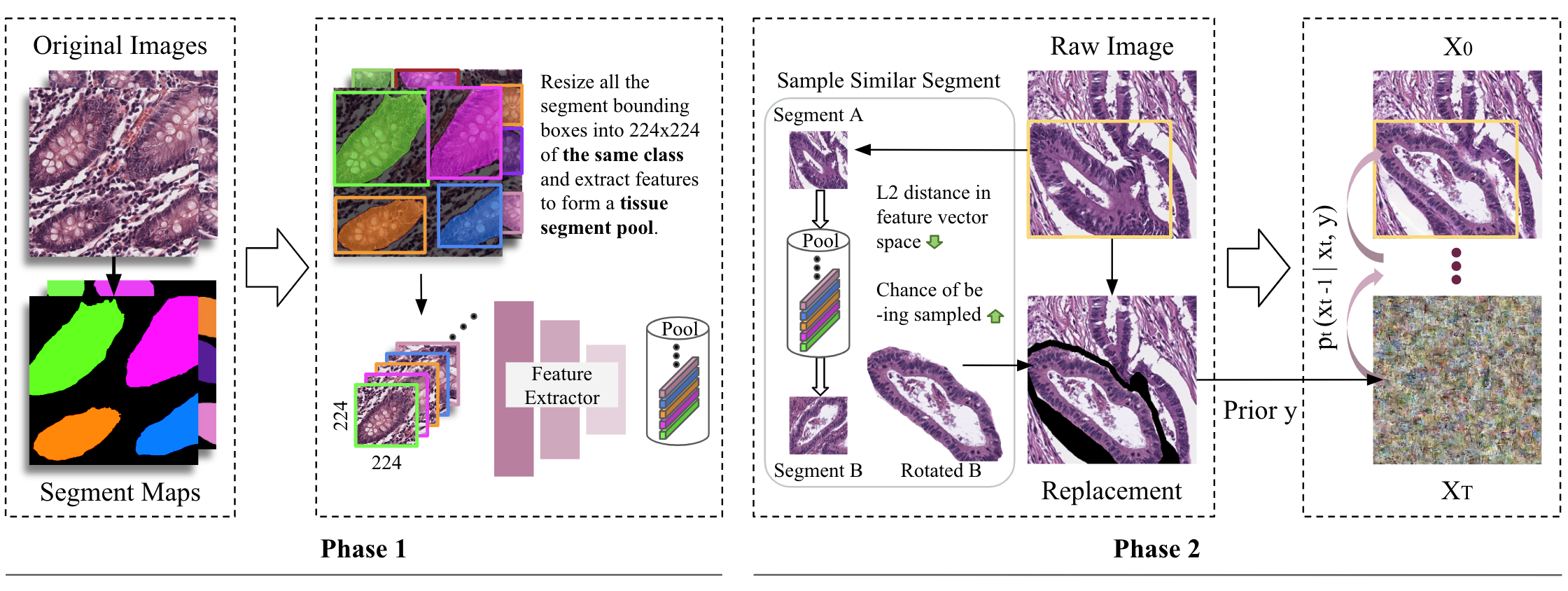}
\caption{The two-phase procedure of USegMix. In phase 1, it first conducts SAM-based unsupervised segmentation to produce the segmentation map for each image, extracts features for each segment, and builds a tissue segment pool. In phase 2, it randomly selects a segment (Segment A) to replace, selects a similar segment (Segment B) from the segment pool, replaces Segment A with Segment B, and conducts inpainting using a pre-trained diffusion model to enhance the realism of the resultant image.} \label{fig1}
\end{figure}

\section{Methods}
USegMix conducts data augmentation in two phases. In the first phase, it performs unsupervised image segmentation to build a tissue segment pool using superpixels and Segment Anything Model (SAM)~\cite{sam}. In the second phase, it mixes and blends segments from different images by adopting a diffusion model via image inpainting. 
Fig.~\ref{fig1} illustrates the procedure of USegMix. 

\subsection{USegMix Phase 1 - Segment Pool Preparation}\label{USegMix1}
\paragraph{SAM-based Unsupervised Segmentation.}
SAM stands as a foundation model trained on over one billion data samples, which endows it with a robust zero-shot capability for segmentation tasks. SAM encodes the inputs of an image and a prompt such as masks, points, or boxes, and returns the decoded resultant masks.

Since SAM is designed in a user-interactive manner, users need to specify regions of interest to conduct segmentation, limiting its usability. 

\begin{figure}
\centering
\includegraphics[width=\linewidth]{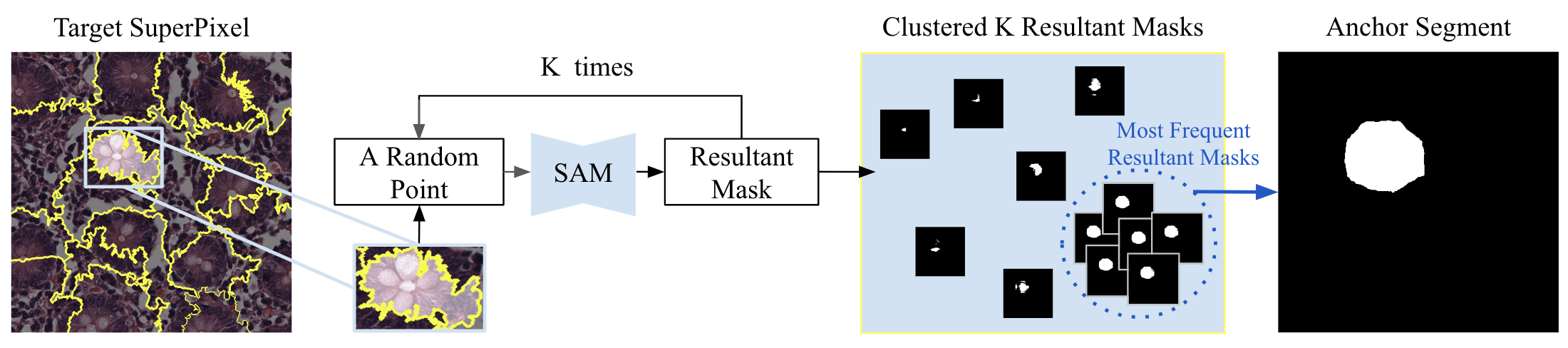}
\caption{The procedure of SAM-based unsupervised segmentation in USegMix phase 1. Given an image, it generates superpixels, selects a target superpixel, samples a random point within the target superpixel, feeds SAM with the random point as a prompt for K times, and collects the resultant masks from SAM. Then, it clusters the resultant K masks based on their similarity and selects the most prevalent mask as the anchor mask to represent the superpixel.} \label{fig2}
\end{figure}

USegMix, however, combines superpixels and SAM to enable an automated generation of segmentation maps, thereby bypassing the need for manual prompts. The segmentation process begins with a SLIC-based superpixel method~\cite{slic} that outlines preliminary segmentation maps $S={s_i, i=1,...,N_s}$ where $s_i$ is the $i$th superpixel and $N_s$ is the number superpixels ($N_s$=30). For a superpixel $s_i$, we randomly sample a point within $s_i$ and use it as a prompt to SAM to generate a segment mask. By repeating it K times with differing random points, we obtain K segment masks, producing $\{m_{i,j}|j=1,...,K\}$ where $m_{i,j}$ is the segment mask for the $j$th random point (K=15). For each $m_{i,j}$, we compute an identity vector $(X(m_{i,j}), Y(m_{i,j}), \sqrt{CNT(m_{i,j})})$ where $X(.)$ and $Y(.)$ denote the center coordinates and $CNT(.)$ is the size of the segment mask. Using the identity vectors and L2 distance, we group the segment masks into a number of clusters, choose the largest cluster, and identify the most frequent segment mask as the anchor segment $\alpha_i$. After iterating this process across all superpixels, we remove the duplicate segments and obtain a refined segmentation map $\cup_{i=1} \alpha_i$ in an automated manner. The segmentation procedure is illustrated in Fig.~\ref{fig2}.

\paragraph{Feature Representation.} 
For each image patch of class $c$, each anchor segment $\alpha_i$ is cropped using its bounding box, resized into the size of 224x224, and fed into a feature extractor which collected around 15 million unlabeled patches cropped from WSIs in TCGA and PAIP for the self-supervised pretraining(CTransPath~\cite{ctranspath}), producing a feature vector. Then, we collect the feature vectors within each histologic category and apply Principal Component Analysis (PCA)~\cite{pca} to obtain the final feature vector $f_i \in \mathbb{R}^p$ where $p$ is 128. Using the anchor segments and the corresponding feature vectors, we construct the segment pool for each category $c$ as follows: $\mathcal{P}^{c} = \{(\alpha_i, f_i)\}_{i=1}^{N_c}$ where $N_c$ is the size of the pool and $c$ denotes the class label. 

\subsection{USegMix Phase 2 - Augmentation by Segment Replacement}
\paragraph{Segment Replacement.}\label{seg_rep}
Suppose that we are given a target image with a number of segments from a histologic category $c$. In the target image, we randomly select a target segment $\alpha^t$ for replacement. To identify the proper replacement $\alpha^r$ from $\mathcal{P}^{c}$ that is semantically similar to $\alpha^t$, we devise a probabilistic sampling method. The method first computes the probability distribution of the segments in $\mathcal{P}^{c}$ given $\alpha^t$:

\begin{equation}
    p({\alpha_j | \alpha^t}) = \frac{e^{-w_j \times D(\alpha^t,\alpha_j)}}{\sum_{k=1}^{N_c} e^{-w_k \times D(\alpha^t,\alpha_k)}}
\end{equation}
where $\alpha_j$ is the $j$th segment in $\mathcal{P}^{c}$, $w_j$ is the weight associated with $\alpha_j$, $D(\alpha^t,\alpha_j) = ||f^t - f_j||_2$, $f^t$ is the feature vector of $\alpha^t$, and $||.||_2$ is the L2 distance. 
Then, we select $\alpha^r \sim p({\alpha | \alpha_i^t})$. 
$w_j$ is initially set to 1 and increased by one as $\alpha_j$ is selected as the replacement to penalize $\alpha_j$ and balance the segment selection. Finally, we eliminate $\alpha^t$ from the target image and paste $\alpha^r$ into the corresponding region by applying the affine transformation.

\paragraph{Diffusion Model Inpainting.}

Due to the difference in the size and shape between $\alpha^t$ and $\alpha^r$ and the boundary effect, the simple replacement results in artifacts. To generate non-artifact images, we utilize a pre-trained latent diffusion model (LDM)~\cite{dm,ddpm,sde,smld,ldm} as follows. Given an image $\mathbf{x} \in \mathbb{R}^{H\times W\times 3}$, an encoder $E$ produces the latent representation $\mathbf{z} = E(\mathbf{x})$, a decoder $D$ reconstructs $\mathbf{x}$ by $\tilde{\mathbf{x}} = D(\mathbf{z}) = D(E(\mathbf{x}))$. For $t \in [0, T]$, using stochastic differential equations (SDEs), the forward continuous diffusion process in low-dimensional latent space can be defined as:
\begin{equation}
d\mathbf{z} = \mathbf{f}(\mathbf{z}, t)dt + g(t)d\mathbf{w}
\end{equation} 
where $\mathbf{f}(.,t)$ is the drift coefficient and $g(t)$ is the diffusion coefficient, and $\mathbf{w}$ is the standard Brownian motion. 
Following~\cite{sde}, the coupled reverse SDE conditioned on $\mathbf{y}$ is given by:
\begin{equation}
d\mathbf{z} = \left[ \mathbf{f}(\mathbf{z}, t) - g(t)^2 \nabla_{\mathbf{z}} \log p_t(\mathbf{z}|\mathbf{y}) \right]dt + g(t) \, d\mathbf{w}
\label{eq:sample}
\end{equation}
where $dt$ is an infinitesimal negative time step and $d\mathbf{w}$ is the standard Brownian motion running backward in time.
To find the score function $\nabla_z \log p_t(\mathbf{z})$, we can train an unconditional LDM $s_{\theta}(\mathbf{z}, t)$ using the following objective~\cite{sde}:
\begin{equation}
\underset{\theta}{min} \mathbb{E}_{E(x),\mathbf{z}(0)), \mathbf{z}(t))}\left[ \left\| s_{\theta}(\mathbf{z}(t), t) - \nabla_{\mathbf{z}_t} \log p_{0t}(\mathbf{z}(t) | \mathbf{z}(0)) \right\|_{2}^{2}\right]
\
\end{equation}
where $t \sim U(0,T)$ and $U$ is the uniform distribution.
With the pre-trained unconditional LDM, we sample $\mathbf{z}$ from the conditional distribution given a prior $\mathbf{y}$ as (\ref{eq:sample}) where $\mathbf{y}$ is the image regions other than the artifacts from image replacement step in Section \ref{seg_rep}. 

We also apply the Manifold Constraint Gradient (MCG)~\cite{mcg} method in the process of reverse sampling. MCG mitigates the error accumulation during sampling iterations by adding a correction term and thus boosts inpainting performance with a simple implementation. Eventually, we reconstruct the image $\tilde{\mathbf{x}}$ from the latent space by passing the sampled $\mathbf{z}$ through the decoder $D(\mathbf{z})$.

\begin{figure*}[h]
\centering
\includegraphics[width=\linewidth]{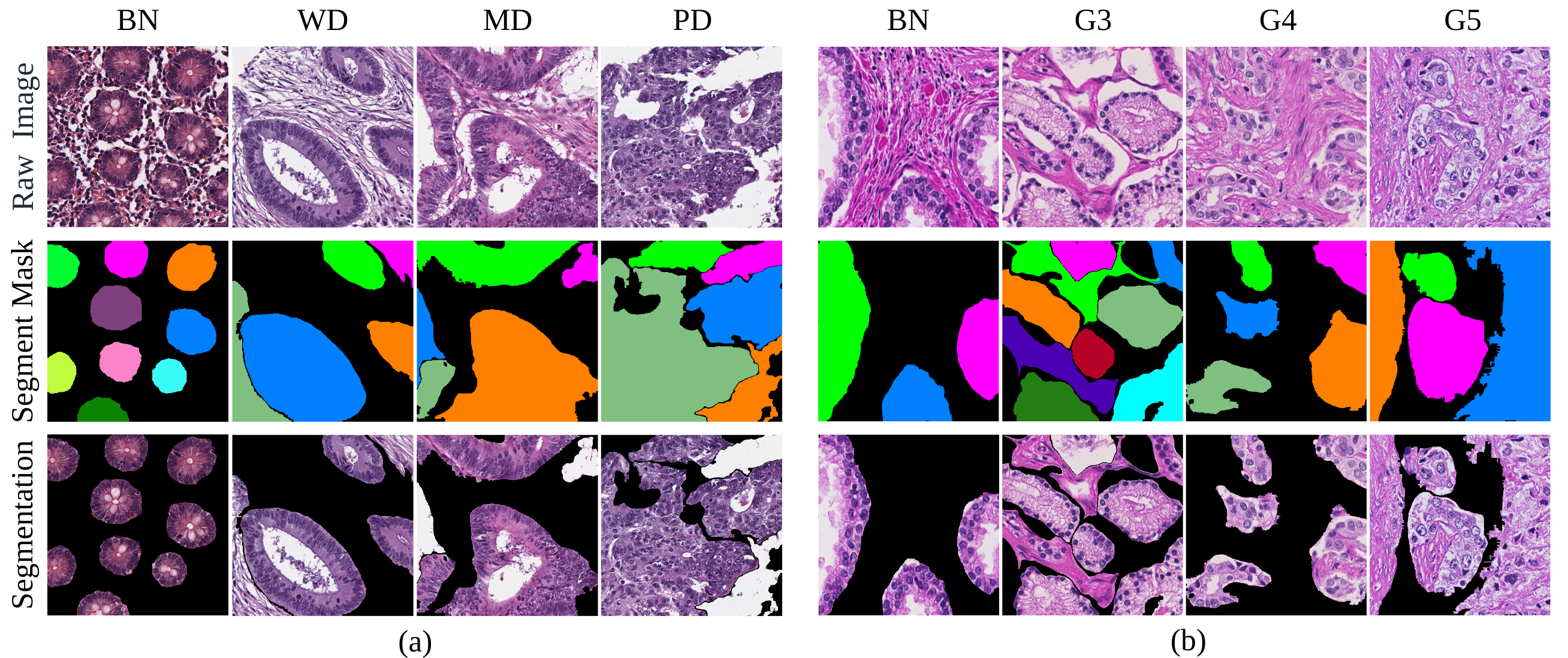}
\caption{Examples of unsupervised segmentation (USegMix Phase 1). (a) Colorectal and (b) prostate tissues.} \label{fig3}
\end{figure*}

\begin{figure*}[h]
\centering
\includegraphics[width=\linewidth]{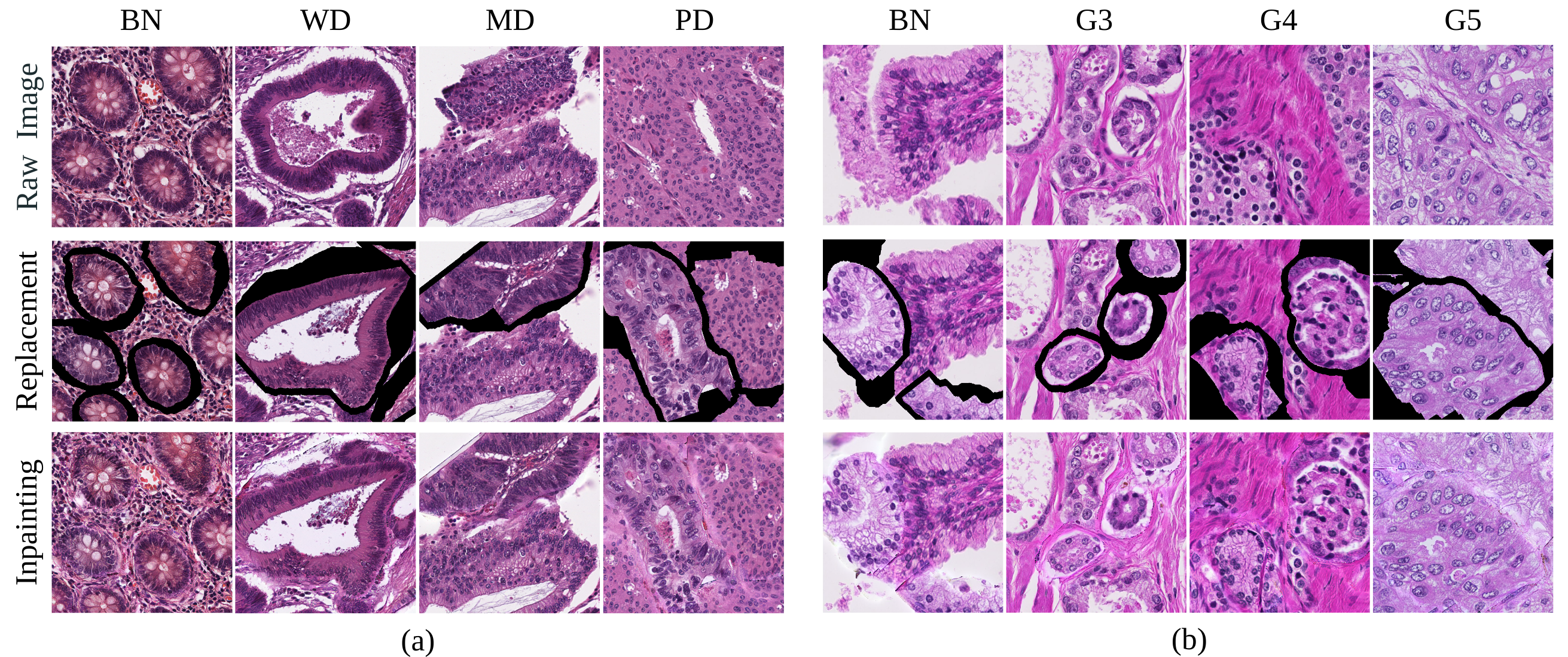}
\caption{Examples of segment replacement and inpainting (USegMix Phase 2). (a) Colorectal and (b) prostate tissues.} \label{fig4}
\end{figure*}

\section{Experiments}
\subsection{Datasets}
\paragraph{Diffusion Model Training Dataset.}
To train an unconditional LDM~\cite{ldm}, we employ comprehensive datasets including seven different tissue types: 1) Colorectal: 10k image patches cropped from 50 whole-slide images (WSIs) of PAIP2021 training set~\cite{paip2021}, 10k image patches obtained from MedFM2023~\cite{medfm2023}, and 8k image patches from \cite{colon_zendo}; 2) Prostate: 10k image patches cropped from 50 WSIs of PAIP2021 training set~\cite{paip2021}; 3) Pancreas: 10k image patches cropped from 50 WSIs of PAIP2021 training set~\cite{paip2021}; 4) Breast: 10k image patches cropped from 150 WSIs of MIDOG2022~\cite{midog2022}; 5) Neuroendocrine tumor: 5k image patches cropped from 55 neuroendocrine tumor WSIs of MIDOG2022~\cite{midog2022}; 6) Melanoma: 5k image patches cropped from 50 WSIs of MIDOG2022~\cite{midog2022}; 7) Gastric: 10k image patches cropped from 98 WSIs~\cite{gastric}.

\paragraph{Cancer Classification Dataset.}
For cancer classification colorectal cancer and prostate cancer datasets are adopted. For colorectal cancer, two publicly available datasets~\cite{colon} are employed. The datasets are categorized into benign (BN), well-differentiated (WD), moderately-differentiated (MD), and poorly-differentiated (PD) tumors. The first dataset includes three sets including a training set of 7027 image patches, a validation set of 1242 image patches, and a test set of 1588 image patches ($Test I$). The image patches are of size 1024$\times$1024. The second dataset $Test II$ contains 110170 image patches of size 1144$\times$1144 pixels.

As for prostate cancer, we acquired public data from Harvard Dataverse (http://dataverse.harvard.edu), collected by the University Hospital of Zurich (USZ). The dataset includes image patches of size 750$\times$750, annotated with four classes such as benign (BN), Gleason 3 (G3), Gleason 4 (G4), and Gleason 5 (G5). The image patches are split into a training set of 15303 image patches, a validation set of 2482 image patches, and a test set of 4237 image patches ($Test$).

\subsection{Implementation Details}

For each target image, we control the new-area ratio (the proportion of the replacement and inpainting areas relative to total image size) to fall within the 30$\%$$\sim$100$\%$ range. 
We train an unconditional LDM~\cite{ldm} for 200 epochs using 78k image patches of 512$\times$512 pixels, loaded with the pre-trained weight of 256$\times$256 LSUN-bedrooms~\cite{lsun}. For image generation, we use 500 DDIM~\cite{ddim} steps.
For each dataset, we synthesize 600 new images for each class.

We adopt four classification models (EfficientNet-B0~\cite{efficientnet}, ResNet-50\cite{resnet}, DenseNet121\cite{densenet}, and CTransPath\cite{ctranspath}), pre-trained on ImageNet, and train them with Adam optimizer (${\beta}_1 = 0.9, {\beta}_2 = 0.999, {\epsilon} = 1.0e^{-8}$) and cosine annealing warm restarts scheduler.
For EfficientNet-B0, ResNet-50, and DenseNet121, we set the initial learning rate to $1.0e^{-3}$, $eta\_min=1.0e^{-5}$, and $T_0 = 25$. The learning rate decreases to $1.0e^{-5}$ after every 25 epochs and restarts from $1.0e^{-3}$. The image patches are resized to 512$\times$512.
As for CTransPath, the image patches are resized to 224$\times$224, and the initial learning rate is set to $2.0e^{-5}$, and $eta\_min=2.0e^{-8}$. 
A batch size is set to 64 for the colorectal and prostate datasets. 50 epochs are used for the colorectal dataset and 100 epochs are employed for the prostate dataset. Traditional data augmentation techniques include random horizontal and vertical flips, Gaussian blurring, Gaussian noise, color change in hue, saturation, and modification of contrast, implemented using Aleju library (https://github.com/aleju/imgaug).
All the models are implemented in PyTorch and run on two NVIDIA RTX A6000 GPUs.

\subsection{Results}

\begin{table*}[!ht]
\caption{Classification results on Colorectal and Prostate Datasets} 
\label{tab:tab1}
\begin{center}    
\resizebox{\linewidth}{!}{
\begin{tabular}{cccccccccccccc} 
\hline
\rule[-1ex]{0pt}{3.5ex} \multirow{2}{*}{Model} & Synthesis  & \multicolumn{4}{c}{Colorectal $Test I$}  & \multicolumn{4}{c}{Colorectal $Test II$} & \multicolumn{4}{c}{Prostate $Test$} \\ 
\cline{3-14}
\rule[-1ex]{0pt}{3.5ex}  & per Class & Acc($\%$) & F1 & Kappa & Avg$\uparrow(\%)$ & Acc($\%$) & F1 & Kappa & Avg$\uparrow(\%)$ & Acc($\%$) & F1 & Kappa & Avg$\uparrow(\%)$ \\
\hline
\rule[-1ex]{0pt}{3.5ex} \multirow{3}{*}{EfficientNet-B0} & No & 85.52 & 0.8062 & 0.9251 & & 79.01 & 0.7118 & 0.8623 & & 68.04 & $\mathbf{0.6339}$ & 0.6043 \\
\cline{2-14}
\rule[-1ex]{0pt}{3.5ex}  & 300 & $\mathbf{87.28}$ & $\mathbf{0.8368}$ & $\mathbf{0.9397}$ & 0.025 & 
79.69 & $\mathbf{0.7387}$ & $\mathbf{0.8845}$ & 0.024 & 
67.64 & 0.6219 & $\mathbf{0.6081}$ & -0.006 \\

\rule[-1ex]{0pt}{3.5ex}  & 600 & 86.34 & 0.8193 & 0.9305 & 0.011 & 
$\mathbf{80.23}$ & 0.7286 & 0.8698 & 0.016 
& $\mathbf{70.57}$ & 0.6290 & 0.6046 & 0.010 \\
\hline
\rule[-1ex]{0pt}{3.5ex} \multirow{3}{*}{ResNet-50} & No & 85.33 & 0.8141 & 0.8767 & & 78.24 & 0.7081 & 0.8381 & & 68.89 & $\mathbf{0.6208}$ & 0.5935 \\
\cline{2-14}
\rule[-1ex]{0pt}{3.5ex}  & 300 & 85.83 & 0.8255 & $\mathbf{0.9295}$ & 0.027 & 
78.28 & $\mathbf{0.7211}$ & $\mathbf{0.8800}$ & 0.023 &
67.93 & 0.6082 & 0.5970 & -0.009\\
\rule[-1ex]{0pt}{3.5ex}  & 600 & $\mathbf{86.40}$ & $\mathbf{0.8279}$ & 0.9041 & 0.020 & 
$\mathbf{78.47}$ & 0.7158 & 0.8229 & -0.001 &
$\mathbf{69.27}$ & 0.6089 & $\mathbf{0.6010}$ & -0.000\\
\hline
\rule[-1ex]{0pt}{3.5ex} \multirow{3}{*}{DenseNet121} & No & 87.03 & $\mathbf{0.8360}$ & 0.9382 & & 78.87 & 0.7155 & 0.8696 & & 67.71 & 0.5883 & $\mathbf{0.6126}$ \\
\cline{2-14}
\rule[-1ex]{0pt}{3.5ex}  & 300 & 86.90 & 0.8297 & 0.9335 & -0.005 
& $\mathbf{82.26}$ & $\mathbf{0.7429}$ & $\mathbf{0.8927}$ & 0.036
& $\mathbf{69.48}$ & $\mathbf{0.6152}$ & 0.6054 & -0.020 \\
\rule[-1ex]{0pt}{3.5ex}  & 600 & $\mathbf{87.34}$ & 0.8309 & $\mathbf{0.9401}$& -0.000 
& 81.41 & 0.7423 & 0.8919 & 0.032
& 64.55 & 0.5911 & 0.5374 & -0.055\\
\hline
\rule[-1ex]{0pt}{3.5ex} \multirow{3}{*}{CTransPath} & No & $\mathbf{87.97}$ & $\mathbf{0.8385}$ & $\mathbf{0.9436}$ & & 81.84 & 0.7518 & 0.9027 & & $\mathbf{72.34}$ & 0.6440 & 0.6208 \\
\cline{2-14}
\rule[-1ex]{0pt}{3.5ex}  & 300 & 86.78 & 0.8207 & 0.9380 & -0.014 
& $\mathbf{82.32}$ & $\mathbf{0.7642}$ & $\mathbf{0.9055}$ & 0.008
& 72.03 & 0.6403 & 0.6221 & -0.003 \\
\rule[-1ex]{0pt}{3.5ex}  & 600 & 87.72 & 0.8373 & 0.9423 & -0.002 & 
79.53 & 0.7388 & 0.8944 & -0.018 
& 71.58 & $\mathbf{0.6480}$ & $\mathbf{0.6249}$ & 0.001\\
\hline
\end{tabular}
}
\end{center}
\end{table*}

\paragraph{Image Synthesis.}
Fig.~\ref{fig3} shows the results of unsupervised segmentation using superpixels and SAM, i.e., USegMix Phase 1, for the four exemplary samples from the colorectal and prostate datasets. It is obvious that USegMix was able to identify/segment most of the histologic objects such as glands in an accurate and reliable manner. 
Moreover, Fig.~\ref{fig4} depicts the results of the segment replacement and inpainting (USegMix Phase 2) for the four samples from the colorectal and prostate datasets. These demonstrate that the replacements are similar to the original segments in the target images and the diffusion model-based inpainting are able to handle artifacts and generate realistic pathology images. 

\paragraph{Cancer Classification.}
To evaluate the quality of the USegMix's synthetic images, we conducted two classification tasks using colorectal and prostate datasets, respectively, using four classification models such as EfficientNet-B0, ResNet-50, DenseNet121, and CTransPath. For each model, we performed three experiments using 1) the original data only, 2) the original data and 300 synthetic images per class, and 3) the original data and 600 synthetic images per class for training.

Table \ref{tab:tab1} shows the results of colorectal cancer classification (Colorectal \textit{Test I} and \textit{Test II}). Overall, the usage of the synthetic images, in general, provided a performance gain for the four classification models regardless of the test datasets. CTransPath for Colorectal \textit{Test I} is the only exception where the one using the original data only gave the best performance; however, on \textit{Test II}, the one with 300 synthetic images was superior to others, the performance drop with 600 synthetic may be due to the overfitting to the training dataset with limited diversity (a large proportion of the synthesized image pixels remain the same) or poor inpaintig qualities (when the stain of the images do not align with the replaced segments). 
The results of prostate cancer classification are available in Table \ref{tab:tab1} (Prostate \textit{Test}). Similar to colorectal cancer classification, the models using the synthetic images, in general, achieved better performance than those with the original data only. However, the four classification models were shown to selectively benefit from the synthetic images; for instance, EfficientNet-B0 with 600 synthetic images obtained the best Acc but the one with 300 synthetic images attained the worst Acc.  
Thus, for the two classification tasks, we observed that the effect of the number of synthetic images varied depending on the type of classification models and the test datasets. 

\section{Conclusion}
We propose USegMix, an efficient data augmentation technique designed to generate realistic synthetic images, enhancing the diversity of pathology datasets and improving the accuracy and robustness of downstream tasks. Experimental results on two cancer datasets using four classification models demonstrate USegMix's efficacy in consistently improving classification performance. USegMix is generic and does not require manual inputs and extra training and processing, and thus it can be applied to various tasks in pathology and other imaging modalities. However, the performance of USegMix was sometimes suboptimal, likely due to the incomplete glands in the patch images and segment pool, and we observed that adding 600 synthetic images can not further improve the performance than adding 300. The other limitation is that the diffusion model needs to be fine-tuned if the users' dataset type is not included in the seven organ types of our diffusion model training datasets. The future study will entail the further development of USegMix to overcome such issues and its application to WSI.

\begin{credits}
\subsubsection{\discintname}
There are no competing interests.
\end{credits}

\end{document}